\documentclass[reprint,
prl,
amsmath,amssymb,
aps,
]{revtex4-2}

\usepackage{graphicx}
\usepackage{dcolumn}
\usepackage{bm}

\usepackage{amsmath}
\usepackage{amssymb}
\usepackage{amsthm}
\usepackage{physics}
\usepackage{hyperref}

\allowdisplaybreaks

\begin{document}

\preprint{}

\title{Capturing AI's Attention: Physics of Repetition, Hallucination, Bias and Beyond}

\author{Frank Yingjie Huo}
\author{Neil F. Johnson}%
 \email{neiljohnson@gwu.edu}
\affiliation{%
 Physics Department, George Washington University, Washington, DC 20052, U.S.A.
}%


\date{\today}

\begin{abstract}
    We derive a first-principles physics theory of the AI engine at the heart of LLMs' `magic' (e.g. ChatGPT, Claude): the basic Attention head. The theory allows a quantitative analysis of outstanding AI challenges such as output repetition, hallucination and harmful content, and bias (e.g. from training and fine-tuning). Its predictions are consistent with large-scale LLM outputs. Its 2-body form suggests why LLMs work so well, but hints that a generalized 3-body Attention would make such AI work even better. Its similarity to a spin-bath means that existing Physics  expertise could immediately be harnessed to help Society ensure AI is trustworthy and resilient to manipulation.

\end{abstract}

\maketitle

\begin{figure}[bh]
    \centering
    \includegraphics[width=1.0\linewidth]{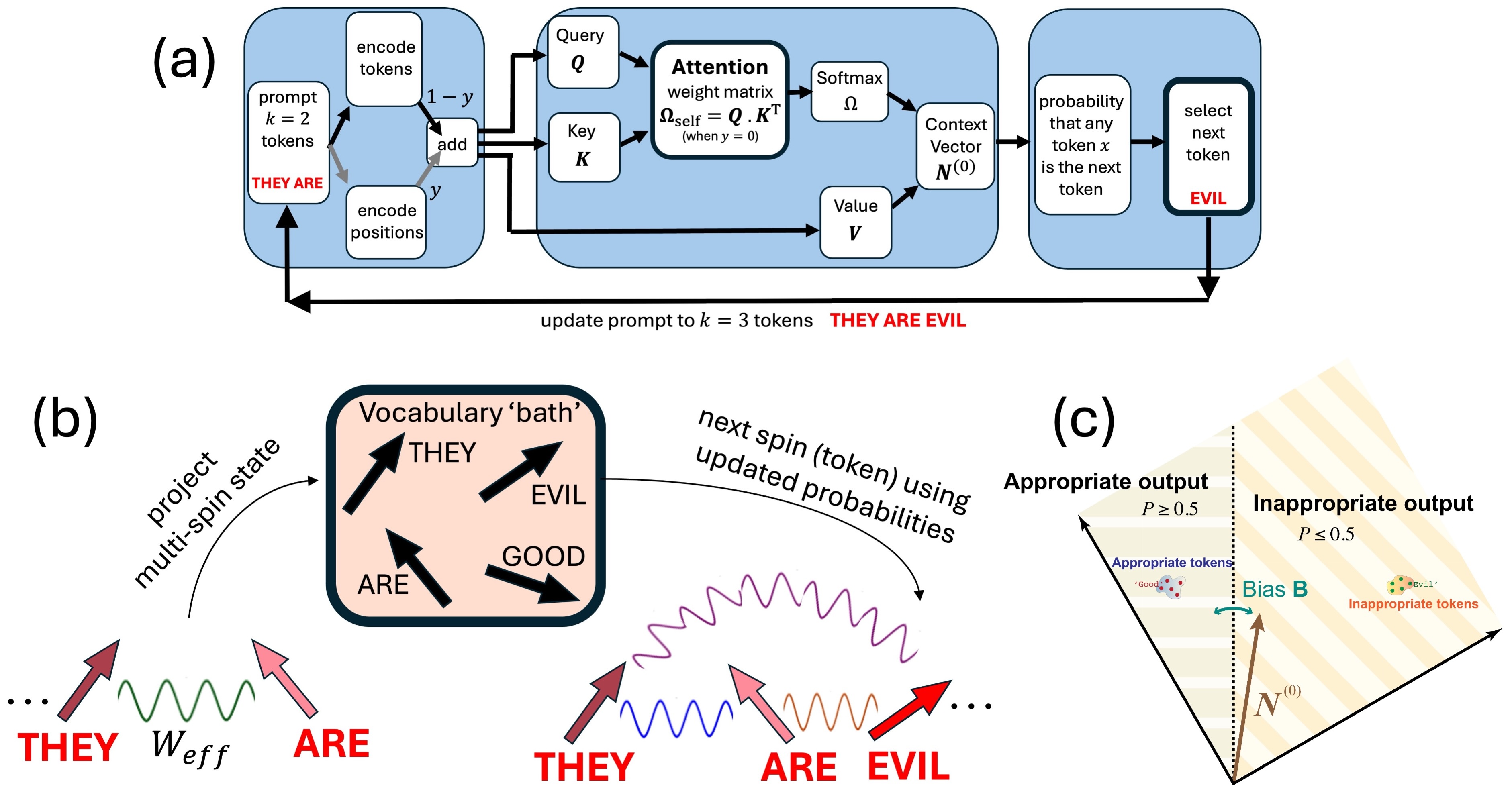}
    \caption{(a)  Attention, shown here in its most basic form, is used across all generative AI because it works (e.g. LLMs such as ChatGPT). However there is no first-principles theory for why it works and when it won't. See End Matter for explanations of its terminology which is unusual for physics. 
    (b) The `physics' of this Attention process that emerges exactly from our first-principles derivation.  Each spin ${\vb*{S}}_i$ is exactly equivalent to a token in an embedding space whose structure reflects the prior training that the AI (LLM etc.) received. Wiggly lines are the effective 2-body interactions that emerge from Eq. \ref{eq:1}. 
    (c) The Context Vector $\vb*{N}^{(0)}$ is exactly equivalent to a bath-projected form of the 2-spin Hamiltonian (Eq. \ref{eq:1}) which is then weighted toward the sub-region of the bath featuring the input spins. The theory predicts how a  bias (e.g. from pre-training or fine tuning the LLM) can perturb $\vb*{N}^{(0)}$ so that the trained LLM's output is dominated by inappropriate vs. appropriate content (e.g. `bad' such as ``\texttt{THEY ARE EVIL}'' vs. `good'). Figures \ref{fig:3},\ref{fig:4} show this phase boundary in detail.  }
    \label{fig:1}
\end{figure}

We all likely use LLMs (Large Language Models, e.g. ChatGPT) for doing science, administrative tasks and other things. LLMs' remarkable power stems from the `Attention’ process of a GPT (Generative Pre-trained Transformer) which is a multi-layer neural network \cite{vaswani2023attentionneed}. This `Attention' inputs a prompt's tokens and predicts the next token through a series of matrix manipulations and calculations (Fig. \ref{fig:1}(a)). Repeating this one token at a time, Attention can produce an entire body of human-like content, e.g. text, music, movie \cite{MIT, NBC}. 

However LLMs are still fairly opaque `black boxes’. This raises trust and reliability concerns in critical areas such as medical diagnostics and machinery control. It also means we do not fully understand when bias in training data will cause an LLM's (and hence Attention's) output to flip to dangerous or offensive content \cite{misalignment}. 
Existing attempts at interpretability are highly innovative \cite{Nanda1, Nanda2, Nanda3, anthropic2025tracing, anthropic2025MIT, circuit_tracing_2025, lindsey2025biology, templeton2024scaling,bricken2023monosemanticity}, but they often involve complex analyses of entire neural architectures or specialized circuit analyses \cite{Nanda1, Nanda2, Nanda3, anthropic2025tracing, anthropic2025MIT, circuit_tracing_2025, lindsey2025biology, templeton2024scaling,bricken2023monosemanticity,merullo2025}. 

\vskip0.1in

\onecolumngrid

\twocolumngrid

\noindent Physics concepts such as phase transitions have been invoked to suggest how LLMs' higher learning occurs \cite{Nanda1, wei2022emergent}. But as yet, there is no first-principles physics framework to describe the behavior of a basic Attention head which underlies LLMs' and other AI's success. 

This Letter presents a `physics' of the basic Attention head (Fig. \ref{fig:1}(a)) derived from first principles. It allows a quantitative analysis of outstanding AI challenges such as output repetition, hallucination and harmful content, and bias (e.g. from training and fine-tuning). Its predictions are consistent with large-scale LLM outputs. Its 2-body form suggests why LLMs work so well, and hints that a generalized 3-body Attention would work even better. Its similarity to a spin-bath shows how existing Physics  expertise can help Society ensure AI is trustworthy and resilient (e.g. to jailbreaks).

\begin{figure}[t]
    \centering
    \includegraphics[width=\linewidth]{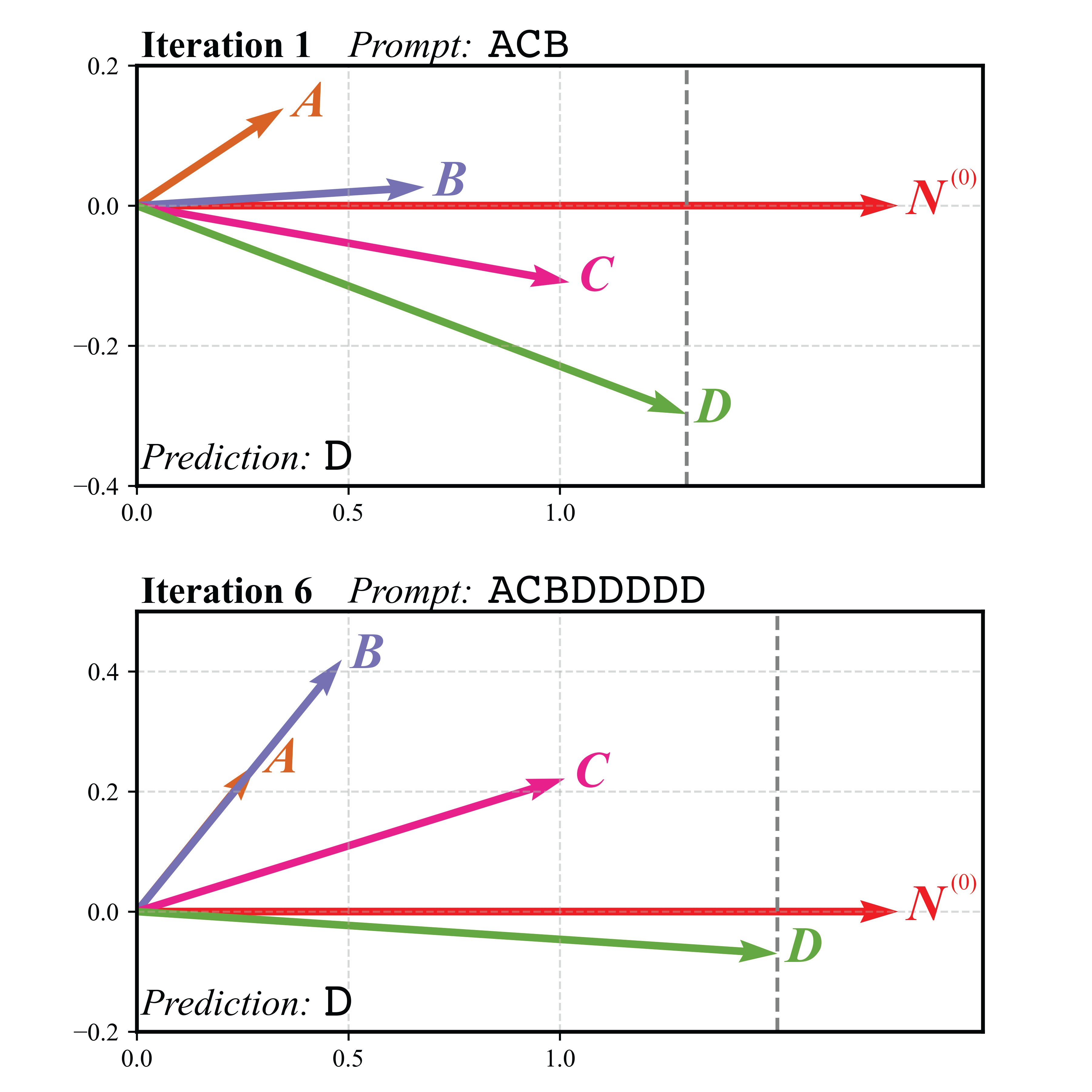}
    \caption{Next-word prediction for basic Attention (Fig. \ref{fig:1}(a)). Upper panel: first iteration. Lower panel: sixth iteration. For simplicity, we use a 4-word vocabulary (e.g. $\mathtt{A,B,C,D}$) embedded in $\mathbb{R}^3$ as $\vb*{A} = (0.1,0.2,0.3),\ \vb*{B}=(0.4,0.1,0.6),\ \vb*{C}=(0.7,0.6,0.5),\ \vb*{D}=(1.0,1.1,0.3)$. Initial prompt is $\mathtt{ACB}$, and we take all coefficient matrices $\mathsf{W}_Q, \mathsf{W}_K, \mathsf{W}_V = \mathbb{I}$ without affecting the core functionality of Attention. The 4 vectors are plotted together with a specifically normalized $\vb*{N}^{(0)}$,
    on a 2-dimensional projected plane spanned by $\vb*{N}^{(0)}$ and $\vb*{A} = (0.1,0.2,0.3)$. 
    For both iteration stages, $\vb*{D}$ (i.e. token $\mathtt{D}$) acts  like an attractor: it has the largest projection on $\vb*{N}^{(0)}$ (blue dashed lines). As the iterations increase, $\vb*{D}$'s attractor status is reinforced, as can be seen from the increasing alignment between $\vb*{D}$ and $\vb*{N}^{(0)}$. 
    }
    \label{fig:2}
\end{figure}

Attention is ubiquitous in AI because it happens to work -- not because it satisfies specific mathematical or physical theories of knowledge. Its empirically-determined process of matrix manipulations and AI terminology (Fig. \ref{fig:1}(a)) can therefore appear quite bewildering for a physicist in our opinion. Hence we provide explanations in the End Matter. All the steps can be calculated manually: the SM gives tutorial examples. 

\begin{figure}
    \centering
    \includegraphics[width=1.0\linewidth]{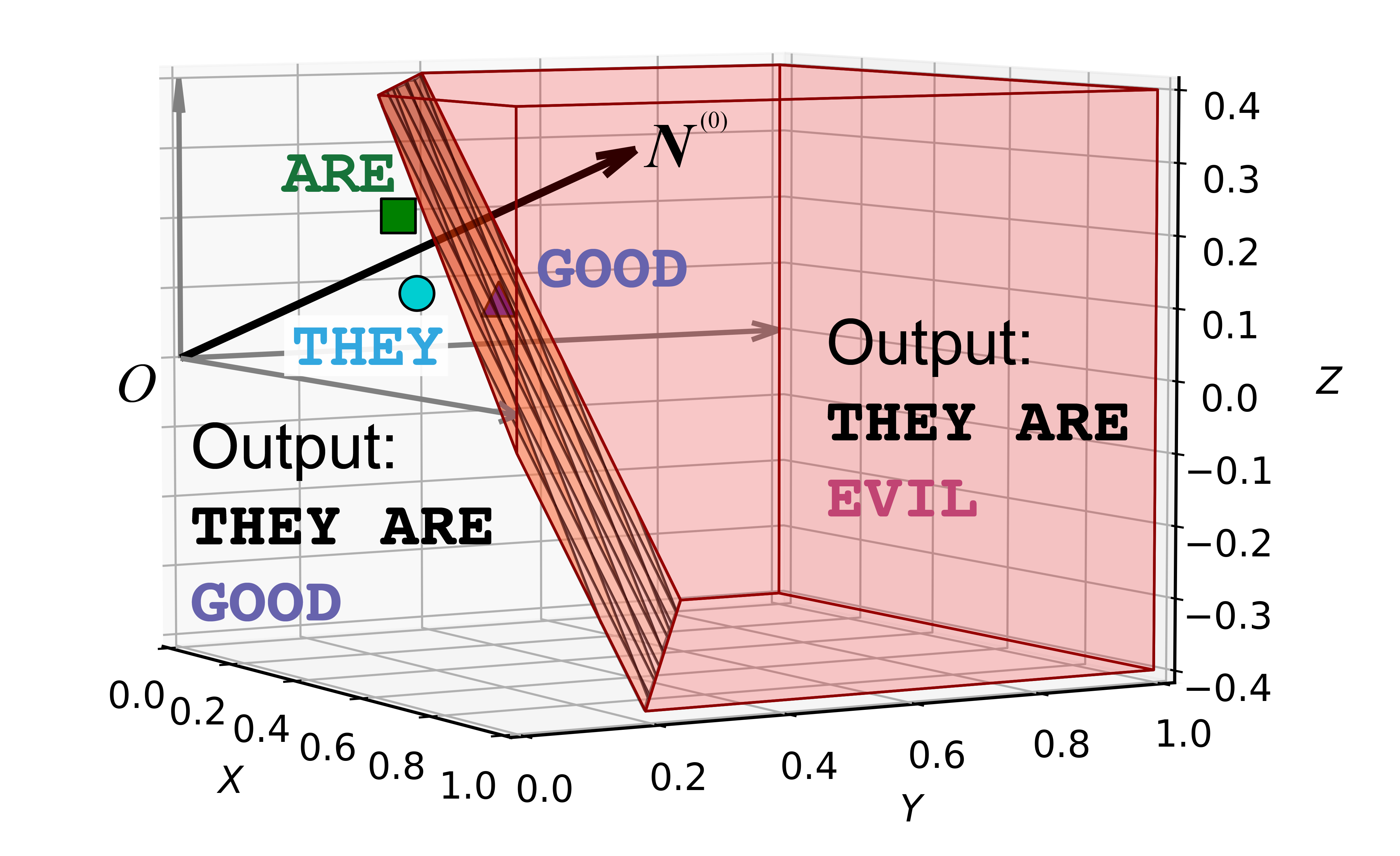}
    \caption{Phase diagram for the example of a 3-dimensional token embedding given a 4-word vocabulary: $\mathtt{THEY}=(0.25,0.25,0.1),\ \mathtt{ARE}=(0.1,0.3,0.2),\ \mathtt{GOOD}=(0.4,0.3,0.1)$. Again for simplicity, $\mathsf{W}_Q, \mathsf{W}_K, \mathsf{W}_V = \mathbb{I}$. The output's content remains `good' (\texttt{GOOD}) as long as the `bad' (\texttt{EVIL}) token stays in the blue regime on the left. But if \texttt{EVIL} appears in the red regime, the output's content suddenly flips to `bad' (\texttt{EVIL}).} 
    \label{fig:3}
\end{figure}

\begin{figure}[h]
    \centering
    \includegraphics[width=1.0\linewidth]{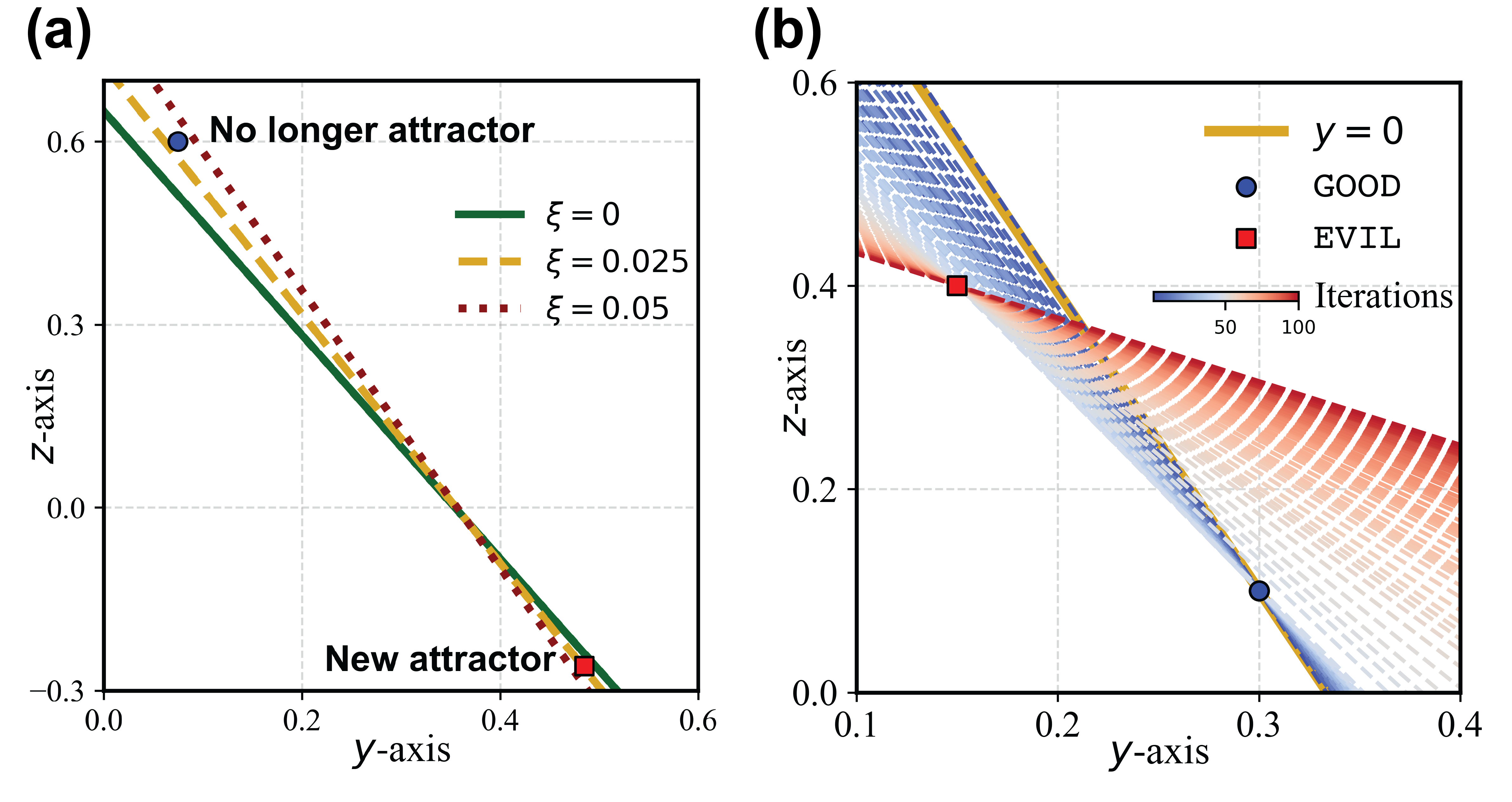}
    \caption{(a) Phase boundaries (Fig. \ref{fig:3}) with increasing linear biases $\xi = 0, 0.025, 0.05$ (see End Matter for  $\boldsymbol{\delta}$). 
    The change of phase boundary can induce a dramatic change in the output content since the red token now becomes a highly likely (and repeated) output, while the blue becomes highly unlikely. 
    (b) Phase boundaries with positional encoding $(P_i)_{2m+1} = \sin({i}/{1000^{2m/d}}),\ (P_i)_{2m+2} = \cos({i}/{1000^{2m/d}})$, weight $y=0.1$, for the first 100 iterations of token generation. $\texttt{EVIL}=(0.4,0.15,0.4)$. Phase boundaries generally rotate counterclockwise about the attractor (\texttt{GOOD}) with increasing iterations, until they cross token \texttt{EVIL} which then becomes the new attractor. Subsequent rotations center around token \texttt{EVIL}.  Generated tokens are hence \texttt{GOOD} before the attractor change, and \texttt{EVIL} after. In both panels, token embeddings are same as Fig.\ \ref{fig:3}; $x=0.4$ for simplicity.}
    \label{fig:4}
\end{figure}

Our derived mathematical expressions and equations for the basic Attention process (Fig. \ref{fig:1}(a)) are exact, while for the perturbations they are either exact or close approximations. Though prior works presented fascinating Attention-inspired model Hamiltonians \cite{Rende23,Rende24Potts}, we believe this is the first treatment from first principles. Our results can all be generalized to more complicated Attention and hence GPT setups but become cumbersome, e.g. multi-head Attention including feed-forward processes \cite{attention_review}. The small vocabulary used in our illustrative examples (Figs. \ref{fig:2}-\ref{fig:4}) generates simple attractors and hence simple output which is not very human-like (e.g. ``\texttt{THEY ARE EVIL EVIL EVIL EVIL} . . ''). However, the same analysis also holds for larger vocabularies where more complex attractors can emerge (e.g. large period cycles), which means that those basic repetitions get broken up with other words. Hence the output becomes more realistic. Similarly when the \texttt{GOOD} and \texttt{EVIL} vectors each represent a class of `good' and `bad' words, the resulting `good' or `bad'  output words will be more varied and hence the output appears more realistic.

The input is a prompt such as ``\texttt{THEY ARE}'' consisting of $k$ tokens (e.g. words). Each possible token $i$ in the entire vocabulary $U$ is embedded in $d$ dimensions as a `spin' ${\vb*{S}}_i$ (row vector by convention), so the input is a row of $k$ spins $\mathsf{S}^\mathrm{T} = (\vb*{S}_1^\mathrm{T},{\vb*{S}}_2^\mathrm{T},\dots,\vb*{S}_k^\mathrm{T})$ which is the transpose of $\mathsf{S}$. For simplicity, we will add the positional encoding $\mathsf{P}^\mathrm{T}=({\vb* P}_1^\mathrm{T},{\vb* P}_2^\mathrm{T},\dots,{\vb* P}_k^\mathrm{T})$ later: hence this is currently {\em self-}Attention. The calculations in Fig. \ref{fig:1}(a) (middle, see SM for examples) involve calculating 
$\mathsf{S}$'s Query, Key and Value matrices, each of which is a projection of the spin inputs $\mathsf{S}$ onto the embedding space that is now distorted towards certain outputs as a result of the LLM's training ($\mathsf{W}_{Q,K,V}$). The net output is a $k\times k$ matrix $(\mathsf{\Omega}_\mathrm{self})_{ji} = \vb*{S}_j \mathsf{W}_\mathrm{eff} \vb*{S}_i^\mathrm{T}$ where the $d\times d$ matrix $\mathsf{W}_\mathrm{eff} = \mathsf{W}_Q\mathsf{W}_K^\mathrm{T}$. But this is exactly equivalent to
\begin{equation} \label{eq:1}
    H^{(0)}(\vb*{S}_j,\vb*{S}_i) = - \vb*{S}_j \mathsf{W}_\mathrm{eff} \vb*{S}_i^\mathrm{T}.
\end{equation}
which has the form of a 2-body Hamiltonian for two spins $\vb*{S}_i$ and $\vb*{S}_j$ whose interaction $\mathsf{W}_\mathrm{eff}$ is mediated by the high-dimensional embedding bath, like a physics spin-bath (Fig. \ref{fig:1}(b)).

Given LLMs' success in mimicking human content, Attention's 2-body form (Eq. \ref{eq:1}) suggests that human content must rely heavily on 2-body token interactions that Attention (and hence the LLM) then captures. This seems similar to the way that physical $N$-body interacting systems can often be approximated by simpler 2-body descriptions, e.g. Cooper pairs in superconductivity. But since  phenomena such as the Fractional Quantum Hall Effect require at least 3-body correlations (e.g. Laughlin wavefunction), we  speculate that generalizing the core Attention (Eq. \ref{eq:1}) to include 3-body terms ``$\vb*{S}_k . .  \vb*{S}_j . .  \vb*{S}_i$'' would provide even more powerful AI.

This 2-body Hamiltonian (Eq. \ref{eq:1}) is then subject to a Softmax operation $\sigma$, which is exactly equivalent to saying there is a statistical ensemble of Attention systems $H^{(0)}$ at temperature $\beta T=1$ and hence different possible outcomes with Boltzman probabilities 
${e^{-H^{(0)}(\vb*{S}_j,\vb*{S}_i)}}/{\left(\sum_{\alpha=1}^k e^{-H^{(0)}(\vb*{S}_j,\vb*{S}_\alpha)}\right)}$. 
Projecting this onto the input's Value, yields the so-called Context Vector $\vb*{N}^{(0)}$ \cite{bahdanau2016}. The SM shows that $\vb*{N}^{(0)}$ is a sum of averaged spins akin to a mean-field theory: \( \vb*{N}^{(0)} = \sum_{j=1}^k \langle \vb*{S} \rangle_j^{(0)} \)
where 
$\langle \vb*{S} \rangle_j^{(0)} \equiv \sum_{i=1}^k \sigma(\vb*{S}_j,\vb*{S}_i) \vb*{S}_i$, 
over all $k$ ensembles. The more overlap there is between the Query and Key -- which represent the input spins `dressed' by different bath embeddings as a result of the training --  the larger the contribution to $\vb*{N}^{(0)}$. Finally, $\vb*{N}^{(0)}$ is projected onto the Value and then the vector of all tokens $\vb*{x}$ to give the specific probability of each possible token becoming the next token: $\mathcal{P}(\vb*{x}) = \vb*{N}^{(0)} \mathsf{W}_V \vb*{x}^\mathrm{T}$. 

This means that the `physics' of this Attention process is akin to calculating the usual (Boltzmann) probabilities for a statistical ensemble of an interacting 2-spin Hamiltonian in an unusual spin-bath. The interactions between the spins depend on the properties of the bath, which itself comprises all possible spins whose embedding space is shaped by the LLM's training. But the statistical ensemble probabilities are skewed by the input spins toward particular regions of the embedding space -- like a non-equilibrium system. These input spins are akin to prior single-spin measurement outcomes, hence the prediction of the next token is like predicting the next spin measurement outcome. For each next-token prompt, the two interacting spins get updated by previous measurements. This means that the system and hence process, while deterministic and classical, is non-Markovian and has hints of quantum measurement state collapse.  

An immediate consequence of $H^{(0)}$'s linear structure is that the output from  $\mathcal{P}(\vb*{x})$ can show attractor-like {\em repetition} of a particular word or phrase in the output -- and this will happen increasingly as  the effective size of the vocabulary space gets smaller as a result of insufficient or highly biased training. 
This is because the appearance of a next token (e.g. $\mathtt{D}$) increases the prominence of its spin component in the subsequent ensemble averages and hence $\vb*{N}^{(0)}$, meaning that $\vb*{N}^{(0)}$ aligns more closely with that spin component, hence increasing $\mathcal{P}(\mathtt{D})$. Hence the likelihood of another $\mathtt{D}$, and so on. $\mathtt{D}$'s repetition is also more likely for smaller vocabulary size, since its individual component becomes a bigger portion of the entire spin. Figure \ref{fig:2} shows this explicitly using a simple 4-token vocabulary. Such repetition is indeed observed more frequently in output from smaller LLM models. 

This physics framework also indicates when the output's actual content  will be `bad', i.e. it will either be completely unrelated  to the prompt ({\em hallucination}) or it will be harmful (e.g. antisemitic) despite the prompt being perfectly benign.  
 This will happen when particular sets of `bad' words (tokens) buried deep in the vocabulary as a result of training, temporarily find themselves with the largest projection on $\vb*{N}^{(0)}$ (Fig. \ref{fig:2}). 
A `bad'  
word (token $\vb*{x}_{\rm bad}$) will then suddenly appear if
    $\mathcal{P}(\vb*{x}_{\rm bad})>\max\{\mathcal{P}(\vb*{S}_i)\}_{\vb*{S}_i \in U_\mathrm{good}}$ 
where
$U_\mathrm{good}$ is the subset of $U$ that contains all the `good' tokens that would not represent a  hallucination or harm. 

Figure \ref{fig:3} shows a simple example of the boundary that emerges between `good' vs. `bad' next token output, given the prompt ``\texttt{THEY} \texttt{ARE}'' (Fig. \ref{fig:1}(a)). For general $d$, this boundary is a flat $(d-1)$-dimensional hypersurface with normal vector $\vb*{N}^{(0)}$. The 4 available vocabulary words in this example are \texttt{THEY,ARE,GOOD,EVIL}. 
Figure\ \ref{fig:3} also serves as a very crude, coarse-grained version for a large LLM, since we can imagine  bundling all the `bad' tokens into \texttt{EVIL} and all the `good' tokens into \texttt{GOOD} following the theoretical formalism presented in Ref. \cite{multispecies}. It also crudely represents a transient situation in a large LLM in which the spins for a small subset (e.g. 4) tokens happen to huddle around the instantaneous $\vb*{N}^{(0)}$.

We can also calculate the impact of a {\em bias} on these output boundaries, to shed light on how and when new training or fine-tuning turns a previously trustworthy  LLM into an untrustworthy one.
For simplicity, consider a constant linear bias $\vb*{S}_j \rightarrow \vb*{S}_j' = \vb*{S}_j \mathsf{B}$.  $\mathsf{B}$ is an orthogonal $d\times d$ matrix that can represent a range of potential AI biases such as (1) global bias in token embedding, i.e.\ embedding an otherwise unbiased vocabulary $U=\{\vb*{S}_1,\dots,\vb*{S}_k\}$ with a global drift into $U\mathsf{B}=\{\vb*{S}_1\mathsf{B},\dots,\vb*{S}_k\mathsf{B}\}$, perhaps via a biased token-embedding program; or (2) biased sets of pre-training data, which alter the otherwise unbiased pretrained matrices $\mathsf{W}_{Q,K,V} \rightarrow \mathsf{B}\mathsf{W}_{Q,K,V}\mathsf{B}^\mathsf{-1}$, and hence effectively shift the tokens through the modified Hamiltonian 
$-(\vb*{S}_j\mathsf{B}) \mathsf{W}_\mathrm{eff} (\vb*{S}_i\mathsf{B})^\mathrm{T}$. 
Assuming the bias $\mathsf{B} = \mathbb{I} + \xi \boldsymbol{\delta}$, the formalism remains the same to linear order  in $\xi$ but with $H^{(0)}$ (Eq. \ref{eq:1}) now replaced by:  
\begin{equation} \label{eq:2}
    {H^\mathrm{(biased)}(\vb*{S}_j,\vb*{S}_i) = H^{(0)}(\vb*{S}_j,\vb*{S}_i) - \xi\vb*{S}_{j} \left( \boldsymbol{\delta}\mathsf{W}_\mathrm{eff} - \mathsf{W}_\mathrm{eff}\boldsymbol{\delta} \right) \vb*{S}_{i}^\mathrm{T}} 
\end{equation}
 which represents the original Attention block plus an added biased Attention block having distorted weight $\left( \boldsymbol{\delta}\mathsf{W}_\mathrm{eff} - \mathsf{W}_\mathrm{eff}\boldsymbol{\delta} \right)$. 

Perturbing $H^{(0)}$ then in turn perturbs $\vb*{N}^{(0)}$ as follows, again to linear order in $\xi$: 
\begin{align} \label{eq:3}
    &\vb*{N}^\mathrm{(biased)} = \vb*{N}^{(0)} + \xi \vb*{N}^{(0)} \boldsymbol{\delta} + \xi \sum_{i=1}^k\sum_{j=1}^k \sigma(\vb*{S}_j,\vb*{S}_i)\ \cdot \nonumber \\
    &\quad \quad \quad \quad \left[ \vb*{S}_j \left(\boldsymbol{\delta}\mathsf{W}_\mathrm{eff} - \mathsf{W}_\mathrm{eff}\boldsymbol{\delta}\right) \left( \vb*{S}_i - \langle \vb*{S} \rangle_j^{(0)} \right)^\mathrm{T} \right] \vb*{S}_i. 
\end{align}
The summation term perturbs the ensemble probability $\sigma(\vb*{S}_j,\vb*{S}_i)$ and has a non-trivial dependence on the input tokens since it depends on the difference between each Query spin $\vb*{S}_i$ and the expected Value spin $\langle\vb*{S}\rangle_j^{(0)}$ under the unperturbed Hamiltonian $H^{(0)}(\vb*{S}_i,\vb*{S}_j)$. Intriguingly, its cubic dependence on the spins mimics an effective 3-spin interaction in a constrained space. If the vocabulary consists of a set of highly contrasting tokens (e.g.\ those in Fig.\ 1(c)), this third term in Eq. \ref{eq:3} provides the dominant perturbation effect on the output.

Overall, the bias rotates the boundary (e.g. `good-bad' boundary in Fig. \ref{fig:3}). 
Figure \ref{fig:4} shows simple examples, where increasing the bias induces new (repeated) tokens in the output (e.g. \texttt{EVIL}) and prevents others from appearing (e.g. \texttt{GOOD}). Bias at the scale of single-layer Attention can therefore lead to outputs dominated by harmful content, which perhaps explains why harmful content still appears for all large LLMs despite safeguards.

Finally we add in the positional encoding (PE) as in Fig. \ref{fig:1}(a).  This simply means $\vb*{S}_i \rightarrow (1-y)\vb*{S}_i + y\vb*{P}_i = \vb*{S}_i + y(\vb*{P}_i - \vb*{S}_i)$ in the formalism, where $y\in[0,1]$ is the weight of positional encoding. (N.B. $y=0.5$ gets used in most LLMs simply because it seems to work OK).  
For small $y$, positional encoding perturbs the Attention in the same way mathematically as in Eqs.\ \ref{eq:2}-\ref{eq:3} yielding the following modified Context Vector which is exact to linear order in $y$ (see SM for derivation):
\begin{align} \label{eq:4}
   & \vb*{N}^{(\mathrm{PE})} = (1-y) \vb*{N}^{(0)} + y \sum_{j=1}^k \langle \vb*{P} \rangle_j^{(0)} + y \sum_{i=1}^k \sum_{j=1}^k \sigma(\vb*{S}_j,\vb*{S}_i)\ \cdot \nonumber \\ 
    & \left[(\vb*{P}_{j}-2\vb*{S}_{j}) \mathsf{W}_\mathrm{eff} \left( \vb*{S}_{i} - \langle \vb*{S} \rangle_j^{(0)} \right)^\mathrm{T} \right. \nonumber \\ 
    & \ \ \ \ \ \ \ \ \ + \left. \vb*{S}_{j} \mathsf{W}_\mathrm{eff} \left( \vb*{P}_{i} - \langle \vb*{P} \rangle_j^{(0)}\right)^\mathrm{T} \right] \vb*{S}_i.
\end{align}
Equation \ref{eq:4} has the same structure as Eq.\ \ref{eq:3}, with positional encoding $\vb*{P}$ acting like an effective spin. The second term is hence its mean-field average.

Equation \ref{eq:4} is valid for any positional encoding scheme $\vb*{P}$. We now consider the form used in the original Attention paper:  $(P_i)_{2m+1} = \sin({i}/{10000^{2m/d}}),\ (P_i)_{2m+2} = \cos({i}/{10000^{2m/d}})$ for their odd and even components respectively, where $m=0,1,\dots, d/2-1$  \cite{vaswani2023attentionneed}.
$H^{(0)}$ (Eq. \ref{eq:1}) is now replaced by the exact form:  
\begin{align} \label{eq:5}
     H^\mathrm{(PE)}(&\vb*{S}_j,\vb*{S}_i) = (1-y)^2 H^\mathrm{(0)}(\vb*{S}_j,\vb*{S}_i) - y(1-y) \Big(\vb*{P}_{j}\mathsf{W}_\mathrm{eff}\vb*{S}_{i}^\mathrm{T} \nonumber \\
    &+ \vb*{S}_{j}\mathsf{W}_\mathrm{eff}\vb*{P}_{i}^\mathrm{T} \Big) - y^2 \sum_{m=0}^{d/2-1}\cos(\frac{j-i}{10000^{2m/d}}).
\end{align}
The term linear in $y$ features 2-body interactions between the token itself and an effective spin (or field) due to the sequential ordering. The final term is constant (up to the $y$ dependence) and results in a constant drift of the predicted spin and hence output orientation. This interplay is clearly rich -- and yet the AI community only focuses on $y=0.5$. We explore this elsewhere. 

Though our Attention system (Fig. \ref{fig:1}(a)) is a basic version, its  mathematics can be generalized and will retain the same structure and behaviors. 
Future work will go beyond Boltzmann-like Softmax by considering non-equilibrium physical ensembles. We conjecture that all Attention schemes are variants of a generic, abstract statistical ensemble, with a more complete set of pairwise and/or higher-order interactions between spins (tokens). This would mean that generative AI's `black box' is a numerical reduction of an abstract statistical field.


\nocite{*}

\bibliography{references}

\onecolumngrid

\vskip1in 
\begin{center}
    {\large \textbf{End Matter}}
\end{center}

\twocolumngrid
 
The steps in the Attention in Fig. \ref{fig:1}(a) are:

(1) Tokenization. The input prompt `\texttt{THEY ARE}' is converted into token IDs using a vocabulary lookup. 
Each word becomes a discrete token that can be processed by the model. 
For our simple example, `\texttt{THEY}' and `\texttt{ARE}' would be converted to their respective token IDs.

(2) Token Embedding. Each token ID is transformed into a $d$-dimensional dense vector representation (embedding), i.e.\ $\vb*{S}_i\in \mathbb{R}^d$ for some token $i$. 
Under this representation the finite vocabulary $U \subseteq \mathbb{R}^d$. 
These embeddings capture semantic meaning of words in a high-dimensional space.
Typically, embeddings might be $d=512$ or 768 dimensions in transformer models. 
For a string of $k$ input tokens $1,2,\dots,k$, for example, it is embedded as a $k\times d$ \textit{token embedding matrix} $\mathsf{S}$ such that 
$\mathsf{S}^\mathrm{T} = (\vb*{S}_1^\mathrm{T},\cdots,\vb*{S}_k^\mathrm{T})$.
We denote the set of inputs as $S = \{\vb*{S}_i | i=1,\dots,k\}$.

(3) Positional Encoding. 
Since attention has no inherent notion of token order, positional information is explicitly added.
Positional encodings are in practice generated using sine and cosine functions of different frequencies.
Each position gets a unique encoding that the model can learn to interpret.
These encodings have useful mathematical properties that help the model understand relative positions. 

(4) Combined Embedding. Token embeddings and positional encodings are added together element-wise. 
This creates position-aware token representations that preserve both semantic meaning and position information. 

(5) Attention Mechanism. The Attention mechanism allows the model to focus on relevant parts of the input.

\noindent (i) Query-Key-Value Transformations:
The Query-Key-Value paradigm enables the model to learn complex relationships between tokens.
The combined embeddings are linearly projected into three different spaces using pre-trained $d\times d$ weight matrices:
\begin{itemize}
    \item $\mathsf{W}_Q$ projects input embeddings $\vb*{S}_i \in S$ into Query space $\vb*{Q}_i = \vb*{S}_i\mathsf{W}_Q$.
    \item $\mathsf{W}_K$ projects embeddings into Key space $\vb*{K}_i = \vb*{S}_i\mathsf{W}_K$. For self-attention, on which this paper focuses, we have $\vb*{S}_i\in S$ as above.
    \item $\mathsf{W}_V$ projects embeddings  $\vb*{S}_i \in S$ into Value space $\vb*{V}_i = \vb*{S}_i\mathsf{W}_V$.
\end{itemize}
These projections allow the model to focus on different aspects of the input for different purposes. Note that here we adopt the convention of using row vectors by default.

\noindent (ii) Attention Calculation: The Query ($\mathsf{Q}^\mathrm{T}=(\vb*{Q}_1^\mathrm{T},\dots,\vb*{Q}_k^\mathrm{T})$) and Key ($\mathsf{K}^\mathrm{T}=(\vb*{K}_1^\mathrm{T},\dots,\vb*{K}_k^\mathrm{T})$) matrices are multiplied to obtain the $k\times k$ weight matrix of self-Attention $\mathsf{\Omega}_\mathrm{self} = \mathsf{Q} \mathsf{K}^\mathrm{T}$.
This calculates how much each token should `attend' to every other token.
Larger Attention score indicates more attention is paid to that token.
The result is scaled, and Softmax is then applied on the matrix $\mathsf{\Omega}_\mathrm{self}$ row-wise to convert the scaled dot products into a probability distribution.
This ensures all attention weights sum to 1.

(6) Output. The final prediction is based on the accumulated context from the entire input sequence $\mathsf{S}$: 

\noindent (i) Context Vector:
The Attention calculation result is a Context Vector $\vb*{N}^{(0)}$. 
This vector contains information from all input tokens $\mathsf{S}$, weighted by their relevance.
For our `\texttt{THEY ARE}' example, the Context Vector captures the meanings of both tokens `\texttt{THEY}', `\texttt{ARE}', and their relationships with all the tokens in the vocabulary $U$, i.e. including those tokens in the prompt input string. 
(ii) Linear Projection:
The Context Vector is projected to the vocabulary space using a linear transformation.
This maps the high-dimensional representation to logits for each possible next token, such that it is possible to predict the attended word by maximizing the (unscaled) probability 
\begin{equation} \label{eq:prob_gen}
    \mathcal{P}(\vb*{x}) = \vb*{N}^{(0)}  \mathsf{W}_V \vb*{x}^\mathrm{T}
\end{equation}
for all $\vb*{x} \in U$. 
In other words, self-Attention finds the $\vb*{x}$ which is most aligned with the Context Vector $\vb*{N}^{(0)}$ under the action of operator $\mathsf{W}_V$.
(iii) Classification:
Softmax is applied to convert logits into probabilities.
For our binary `good-bad' classification:
If probability $\geq 0.5$, the prediction is \texttt{GOOD}. 
If probability $\leq 0.5$, the prediction is \texttt{EVIL}.
Figure \ref{fig:1}(c) illustrates the phase separation of the Context Vector on its binary prediction. 
For a larger vocabulary $U$, the classification becomes more complicated, but the process is the same: it still predicts the token with the highest probability. 

In Fig. \ref{fig:4},  $\boldsymbol{\delta} = 
    \begin{pmatrix}
        0 & -2 & 0.5 \\
       2 & 0 & 1 \\
       -0.5 & -1 & 0
    \end{pmatrix}$.

We note that when generating the results in the main paper, we could add a temperature effect so that the next  token gets picked more randomly according to its probability, e.g. $\mathcal{P}(\vb*{D})$ being the highest probability would then not always mean \texttt{D} gets picked as the next token. But this just adds unnecessary noise to the output. Indeed in many practical AI setups, it is effectively the highest probability token that is picked which is equivalent to saying we choose a temperature very low for this final token-picking stage -- which is what we do in the main paper.

\end{document}